\documentclass[10pt,twocolumn,letterpaper]{article}

\usepackage[pagenumbers]{cvpr} 

\usepackage{graphicx}
\usepackage{amsmath}
\usepackage{amssymb}
\usepackage{booktabs}

\usepackage[pagebackref,breaklinks,colorlinks]{hyperref}

\usepackage[capitalize]{cleveref}
\crefname{section}{Sec.}{Secs.}
\Crefname{section}{Section}{Sections}
\Crefname{table}{Table}{Tables}
\crefname{table}{Tab.}{Tabs.}

\def\cvprPaperID{55} 
\def\confName{CVPR}
\def\confYear{2022}

\usepackage{xcolor}
\usepackage{colortbl}
\definecolor{colorA}{RGB}{76, 114, 176}  
\definecolor{colorB}{RGB}{221, 132, 82}  
\definecolor{colorC}{RGB}{85, 168, 104}  
\definecolor{colorD}{RGB}{196, 78, 82}  
\definecolor{colorE}{RGB}{129, 114, 179}  
\definecolor{colorF}{RGB}{147, 120, 96}  
\definecolor{Gray}{gray}{0.85}

\begin{document}

\title{Fair Comparison between Efficient Attentions}
\author{Jiuk Hong, Chaehyeon Lee, Soyoun Bang and Heechul Jung\\
Kyungpook National University\\
{\tt\small \{hong4497, 123456ccdd, sphzbsy, heechul\}@knu.ac.kr}\\
{\small \url{https://github.com/CreamNuts/Fair-Comparison-between-Efficient-Attentions}
}}

\maketitle

\begin{abstract}
   Transformers have been successfully used in various fields and are becoming the standard tools in computer vision. However, self-attention, a core component of transformers, has a quadratic complexity problem, which limits the use of transformers in various vision tasks that require dense prediction. Many studies aiming at solving this problem have been reported proposed. However, no comparative study of these methods using the same scale has been reported due to different model configurations, training schemes, and new methods. In our paper, we validate these efficient attention models on the ImageNet1K classification task by changing only the attention operation and examining which efficient attention is better.
\end{abstract}

\section{Introduction}
\label{sec:intro}

Transformer~\cite{vaswani2017attention} models have received widespread attention due to their effectiveness in various fields. The transformer consists of a self-attention operation, which has a quadratic complexity of time and memory proportional to the number of input tokens. In natural language processing (NLP), which was the first field to admit the possibility of transformers, various works~\cite{wang2020linformer, shen2021efficient, wu2021fastformer, zaheer2020big, tay2020efficientsurvey} to improve quadratic complexity problems for handling documents consisting of many words have been reported. Such efforts have continued in computer vision~\cite{choromanski2021rethinking, ali2021xcit}. This is because the vision transformer, which creates tokens by dividing an image into patches, has an inverse relationship between patch size and performance~\cite{dosovitskiy2021an}, but quadratic complexity makes it difficult to handle computations when reducing patch sizes.

Previous studies on using efficient self-attention to improve from quadratic to linear complexity have never been compared in the same environment because of several differences between both of them. Those studies have different model configurations and training schemes except for their self-attention operations. Thus, comparing only the influence of efficient self-attention is difficult. Furthermore, the performance of some efficient self-attention proposed in NLP has not been evaluated in vision tasks.

In this paper, we conduct a comparison experiment by changing only self-attention operations to efficient self-attentions with linear time complexity. Here, the number of patches increases from 4x to 16x compared with $14\times14$ or $16\times16$ patches, increasing the computations to an infeasible extent despite using efficient attention. Therefore, we employ the pyramid structure of previous studies~\cite{wang2021pyramid, liu2021swin}. 

As a result, we observe that the performance improves as computational complexity increases for $4\times4$ and $7\times7$ patches, regardless of what kind of self-attention is used, and that efficient attentions do not perform better than normal self-attention without additional methods.

\section{Efficient Self-Attentions}
\label{sec:eff}
The original transformer architecture~\cite{vaswani2017attention} and the vision transformer for image classification~\cite{dosovitskiy2021an} both conduct global self-attention, where the relationships between a token and all other tokens are computed. Global self-attention leads to quadratic complexity problems proportional to the number of tokens, impeding the use of many tokens to obtain fine-grained features for dense prediction. 

To solve this problem, efficient self-attentions~\cite{wang2020linformer, shen2021efficient, wu2021fastformer,ali2021xcit,choromanski2021rethinking}, which characterize linear complexity and global token interaction, have been proposed. In this section, we outline the efficient self-attentions used in this paper and indicate the computation complexity of each method. Here, we only consider self-attention operation for representing complexity. \Cref{tab:efficient} summarizes abbreviations and the complexity for each efficient attention used in this paper. 

\vspace{2.5mm}
\noindent{\bf Self-attention}
The original self-attention operation~\cite{vaswani2017attention} computes the dot products of queries with keys, divides each by $\sqrt {d_k}$, and applies a softmax function to obtain the weights on the values. Then, a linear projection using $W\in\mathbb R^{C\times C}$ is performed. However, this part is omitted in this paper for clarity.
\begin{equation}
\text{SA}(\mathbf q, \mathbf k, \mathbf v)=\mathrm{softmax}\left[\frac{\mathbf q \mathbf k^T}{\sqrt {d_k}}\right]\mathbf v,
\label{eq:orig}
\end{equation}
where $\mathbf q \in \mathbb R^{N\times d_q}$,  $\mathbf k \in \mathbb R^{N\times d_k}$, and $\mathbf v \in \mathbb R^{N\times d_v}$ represent queries, keys, and values, respectively. Generally, the dimension of query $d_q$ equals that of key $d_k$. They are calculated using the input sequence $\mathbf x\in \mathbb R^{N\times d}$ by projecting $\mathbf x$ onto three learnable weight matrixes $W_q\in \mathbb R^{d\times d_q}$, $W_k\in \mathbb R^{d\times d_k}$, and $W_v\in \mathbb R^{d\times d_v}$. $\mathrm{softmax}(\cdot)$ denotes the application of the softmax function along the last axis of the matrix. The computation complexity of original self-attention is $O(N^2C)$.

\vspace{2.5mm}
\noindent{\bf Linformer}
This technique~\cite{wang2020linformer} is considered a naive improvement of self-attention. Since the original self-attention has quadratic complexity with respect to the number of tokens, Linformer projects $\mathbf k$ and $\mathbf v$ into $m\times C$ matrixes using learnable parameters $W_{proj}\in \mathbb R^{m\times N}$ and then performs self-attention operations. The linear self-attention $\text{LA}(\cdot)$ is computed as:
\begin{equation}
\text{LA}(\mathbf q, \mathbf k, \mathbf v)=\mathrm{softmax}\left(\frac{\mathbf q [W_{proj}\mathbf k]^T}{\sqrt {d_k}}\right) W_{proj}\mathbf v.
\label{eq:lin}
\end{equation}
The computation complexity is $O(NCm)$. In our experiments, we set the project dimension $m$ as $\frac N 4$.

\vspace{2.5mm}
\noindent{\bf Efficient attention}
To improve the quadratic complexity, The efficient attention~\cite{shen2021efficient} change the order of operations in self-attention. This method has been proposed as the non-local module~\cite{wang2018non} in a convolutional neural network. $\mathrm{EA(\cdot)}$ denotes the efficient attention and is written as:
\begin{equation}
\text{EA}(\mathbf q, \mathbf k, \mathbf v)=\mathrm{softmax}(\mathbf q)\left[ \mathrm{softmax}(\mathbf k^T)\mathbf v\right ].
\label{eq:eff}
\end{equation}
Here, computation complexity is $O(NC^2)$. 

\vspace{2.5mm}
\noindent{\bf Performer}
The performer~\cite{choromanski2021rethinking} is an improved kernel method of efficient attention. It uses a kernel approximating softmax function via a positive orthogonal random feature and is denoted as $\mathrm{PA}(\cdot)$:
\begin{equation}
\text{PA}(\mathbf q, \mathbf k, \mathbf v)=\frac{\psi(\mathbf q)[\psi(\mathbf k)^T\mathbf v]}{\mathrm{diag}(\psi(\mathbf q)[\psi(\mathbf k)^T\mathbb I_N])},
\label{eq:perf}
\end{equation}
where $\psi(\cdot):\mathbb R^{d_q}\rightarrow \mathbb R^{r}_+$ is the kernel using a positive orthogonal random feature and $\mathbb I_n$ is an $N\times N$ identity matrix. The computation complexity of the kernel and the total computation complexity is $O(NCr)$. In our experiment, we set the project dimension $r$ as $\frac{C}{2}$ to ensure that $r$ is smaller than $N$, according to the original work~\cite{choromanski2021rethinking}.

\vspace{2.5mm}
\noindent{\bf XCiT}
The cross-covariance attention~\cite{ali2021xcit} is a transposed version of self-attention, which operates across the feature dimension rather than the token dimension. This change causes implicit communication between tokens and degrades the quality of representation. Therefore, XCiT~\cite{ali2021xcit} introduced a local patch interaction module (LPI) consisting of two convolution layers. However, we evaluated results both with and without LPI, denoted as LPI and XCA. The cross-covariance attention $\text{XCA}(\cdot)$ is given as:
\begin{equation}
\text{XCA}(\mathbf q, \mathbf k, \mathbf v)=\left[\mathrm{softmax}\left(\frac{{\|\mathbf q\|_2}^T \|\mathbf k\|_2}{\tau}\right) \mathbf v^T\right]^T,
\label{eq:xcit}
\end{equation}
where $\tau$ is a learnable temperature scaling parameter to compensate for reducing the representational power due to $l_2$ normalization. For comparison, we set $\tau$ as the fixed value $\frac{N}{2}$ and removed the $l_2$ regularization from the experiment. The computation complexity is $O(NC^2)$.

\vspace{2.5mm}
\noindent{\bf Fastformer}
Instead of modeling the global attention using matrix multiplication, the fastformer~\cite{wu2021fastformer} employs element-wise multiplication. This method transforms each token representation using its global context. For the global context $\mathbf q'\in \mathbb R^C$ and $\mathbf k'\in \mathbb R^C$, learnable parameters $W_{q'}\in \mathbb R^C$ and $W_{k'}\in \mathbb R^C$ are required. The additive attention $\mathrm{AA}(\cdot)$ is given as:
\begin{equation}
\text{AA}(\mathbf q, \mathbf k, \mathbf v)=\mathbf q + [\mathbf{k'}* \mathbf v]W,
\label{eq:fast}
\end{equation}
where $\mathbf {k'}$ is computed using $\mathbf q'$ and $*$ denotes element-wise multiplication. $W$ is the projection parameter we mentioned but omitted in \Cref{eq:orig}. The computation complexity of additive attention is $O(NC)$.

\vspace{2.5mm}
\noindent{\bf Swin Transformer}
In addition to the above efficient self-attentions, several approaches to solving quadratic complexity exist, including the Swin transformer~\cite{liu2021swin}. It regards global interactions as the core problem and uses a nested window attention with normal self-attention~\cite{vaswani2017attention}. Therefore, the window attention is outside the scope of our experiment. However, since the Swin transformer is an important benchmark for vision transformer, we compare its performance in this study. The computation complexity of the nested window attention is $O(NCw^2)$, where $w$ is the size of the window. In experiment, we set $w$ as 7 and 8 when the patch sizes are 4 and 7, respectively.

\begin{figure*}
  \centering
    \includegraphics[width=\linewidth]{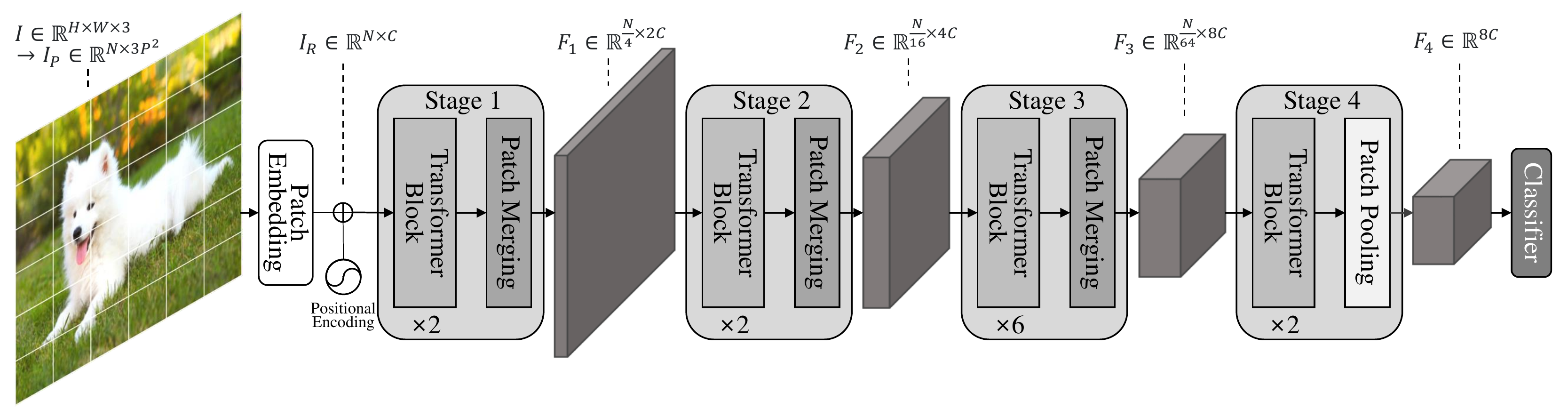}
  \caption{Baseline pyramid architecture of our experiment. We split an image into fixed-size patches, linearly embedded each of them, added positional embedding, and fed the sequence vector $I_R$ to four `Stages'. Next, we performed patch pooling, which is an operation that averages the feature along the patch axis. After computing the last feature $F_4$, we fed the feature into the linear classifier.}
  \label{fig:arch}
  \vspace{-2.5mm}
\end{figure*}

\section{Experiment}
In our experiment, we benchmarked various efficient self-attentions on the ImageNet1K dataset~\cite{deng2009imagenet}, which contains 1.28M training images and 50K validation images from 1,000 classes. For a fair comparison, we trained all models on the training set using a similar configuration and the same training scheme except for self-attention and patch size. We report the top-1 error in the validation set.

\subsection{Baseline architecture}
An overview of the baseline architecture is presented in \Cref{fig:arch}. First, we reshaped the 2D image $I\in \mathbb R^{H\times W\times 3}$ into a sequence of flattened patches $I_P \in \mathbb R^{N\times 3P^2}$, where $P$ and $N$ are the size and number of patches, respectively. If $P$ is 4 and $(H,W)$ is (224, 224), $N$ becomes 3136. Then, we use learned embeddings to convert the patches $I_P$ to the token representation $I_R\in \mathbb R^{N\times C}$ and add positional encoding to $I_R$. Next, this representation was averaged after passing through some `Stage' and input into the linear classifier to obtain logits. Every `Stage' consists of several blocks and we only changed the self-attention layer to efficient self-attentions in the block for a fair comparison.

\vspace{2.5mm}
\noindent{\bf Transformer block}
Blocks of each `Stage' are a multi-head self-attention (MSA) using the efficient attentions, and a feed-forward network (FFN) in a transformer~\cite{vaswani2017attention}. FFN consists of a 2-layer MLP with GELU non-linearity in between. A LayerNorm~\cite{ba2016layer} (LN) layer is applied before each MSA and FFN, and a residual connection is applied after each module.

\vspace{2.5mm}
\noindent{\bf Pyramid structure}
Even if the computation decreases using efficient self-attention, the computation rapidly increases when many tokens are used for attention. Therefore, we employed a pyramid structure to lower the computation to a feasible level. Pyramid structure~\cite{liu2021swin, wang2021pyramid} consists of several `Stages' maintaining the number of tokens. Each `Stage' has several transformer blocks with efficient or normal self-attention. At the end of each `Stage', except the last `Stage', there is a `Patch Merging' layer of which $2\times 2$ neighbor patches are merged to reduce the number of tokens:
\begin{equation}
\mathrm{Patch Merging}(\mathbf x)=\mathrm{Norm}(\mathrm{Reshape}(\mathbf x)W),
\end{equation}
where $\mathbf x\in \mathbb R^{(H\cdot W)\times C}$ represents the input tokens and $W\in \mathbb R^{4C\times2C}$ is a linear projection parameter. $\mathrm{Reshape(\mathbf x)}$ is an operation that reshapes the input sequence $\mathbf x$ to $(\frac{H}{2}\cdot\frac{W}{2})\times 4C$ and $\mathrm{Norm}(\cdot)$ refers to LN.

\vspace{2.5mm}
\noindent{\bf Columnar structure}
Another structure that can be used is a columnar structure, which works similarly to ViT~\cite{dosovitskiy2021an}. This is a reference point for showing the difference between the pyramid and column structures when each structure has a similar amount of computation. Therefore, we do not use efficient attention with a columnar structure. Here, we trained the columnar structure with normal self-attention and set patch sizes to 14 and 16 due to the large computation.

\vspace{2.5mm}
\noindent{\bf Model details}
To make the number of feature channels smaller than that of tokens in each stage, we set the number of feature channels per head to 32. With a pyramid structure, the number of attention layers and heads were set to [2, 2, 6, 2] and [6, 12, 24, 32] along the `Stage'. With a columnar structure, we set the number of heads to 12. Next, we adopted a positional encoding as proposed by El-Nouby \etal~\cite{ali2021xcit}. This method first produces an encoding in an intermediate 64-dimensional space before projecting it to the feature dimension space of the transformer. 

\begin{figure*}[!t]
    \centering
    \includegraphics[width=0.8\linewidth]{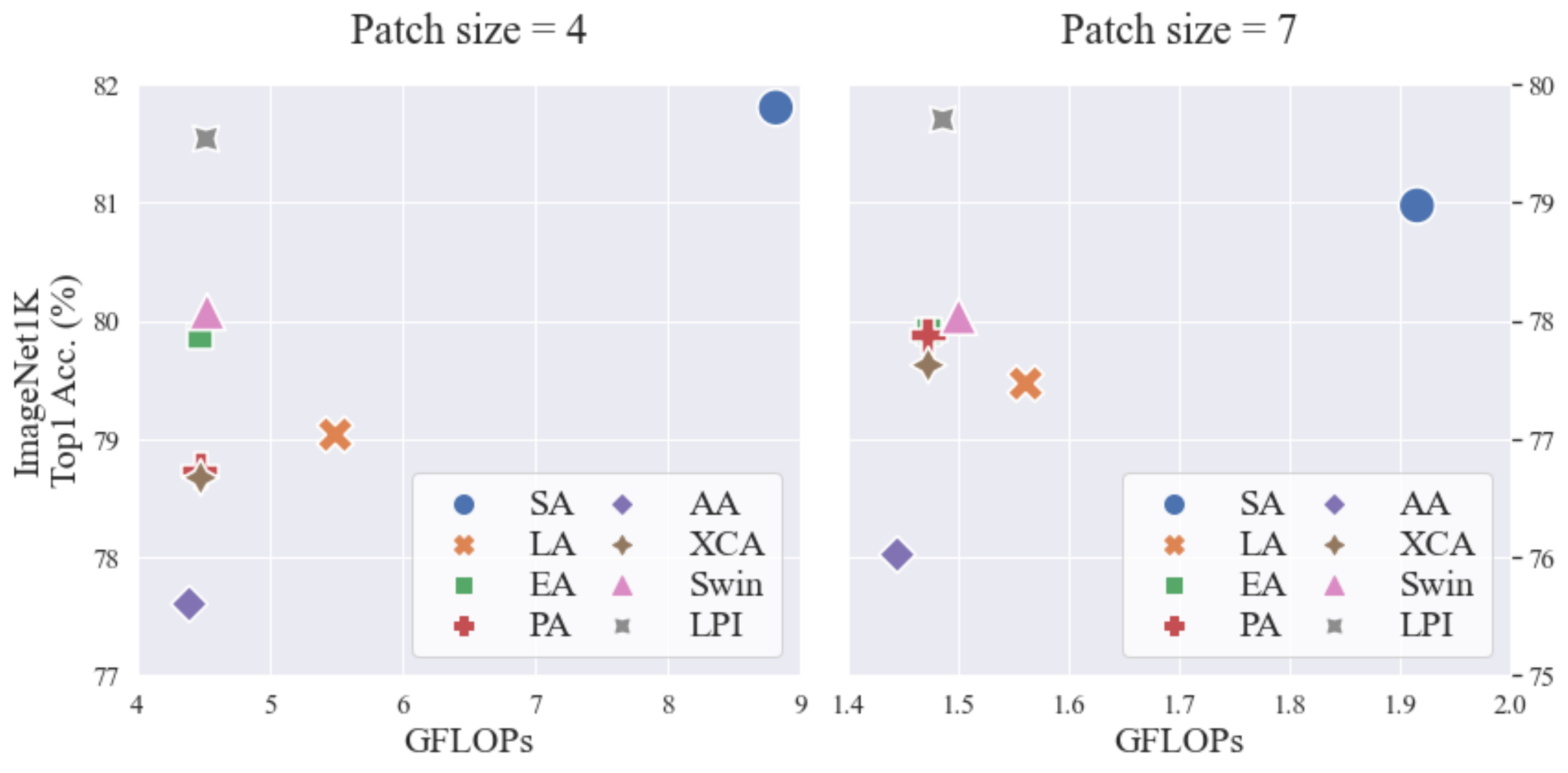}
    \caption{ImageNet1K top 1 accuracy versus GFLOPs under different patch sizes. All models were trained at same scheme.}
    \label{fig:result}
    \vspace{-2.5mm}
\end{figure*}

\subsection{Training Scheme}
For the training, we employed an AdamW optimizer~\cite{loshchilov2017decoupled} for 310 epochs using a cosine decay learning rate scheduler, 20 epochs of linear warm-up, and 10 epochs of cool down. Furthermore, we used a batch size of 128, an initial learning rate of 0.005, and a weight decay of 0.05. Also, we applied most augmentation and regularization strategies of XCiT~\cite{ali2021xcit}. For a baseline with a patch size of 4, we used a batch size of 32 due to limited memory size. For implementation, we developed our code using the \emph{timm library}~\cite{rw2019timm}.

\subsection{Experimental results}
\begin{table}
\centering
    \resizebox{\linewidth}{!}{
    \begin{tabular}{lc}
        \toprule
        Model Architecture & Complexity per Self-Attention\\
        \midrule
        Transformer (SA)~\cite{vaswani2017attention}&$O(N^2C)$\\
        Linformer (LA)~\cite{wang2020linformer}&$O(NCm)$\\
        Efficient Attention (EA)~\cite{shen2021efficient}&$O(NC^2)$\\
        Performer (PA)~\cite{choromanski2021rethinking}&$O(NCr)$\\
        Fastformer (AA)~\cite{wu2021fastformer}&$O(NC)$\\
        XCiT (XCA)~\cite{ali2021xcit}&$O(NC^2)$\\
        Swin Transformer (Swin)~\cite{liu2021swin}&$O(NCw^2)$\\
        \bottomrule
      \end{tabular}}
      \caption{Abbreviations and complexity for each self-attention used. $m$, $r$, and $w$ are their hyperparameters described in \Cref{sec:eff}.}
      \label{tab:efficient}
      \vspace{-2.5mm}
\end{table}

\begin{table}
\centering
    \resizebox{\linewidth}{!}{
    \begin{tabular}{lccc|c}
        \toprule
        Type of & \#params & FLOPs &FLOPs& Top 1\\
        Attention&(M, millions)& (G, giga)&ratio&Acc. (\%) \\
        \midrule
        \multicolumn{5}{c}{Baseline}\\
        \midrule
        SA-4~\cite{vaswani2017attention} & 28.27 & 8.821 & 1 & 81.80\\
        \rowcolor{Gray}
        SA-7~\cite{vaswani2017attention} & 28.28 & 1.915 & 1 & 78.97\\
        \midrule
        \multicolumn{5}{c}{Efficient Attentions}\\
        \midrule
        LA-4~\cite{wang2020linformer} & 30.91 & 5.496 & 0.62 &79.04\textcolor{blue}{{(-2.76)}}\\
        \rowcolor{Gray}
        LA-7~\cite{wang2020linformer} & 28.56 & 1.561 & 0.81 &77.47\textcolor{blue}{{(-1.5)}}\\
        EA-4~\cite{shen2021efficient} & 28.27 & 4.480 & 0.51 &79.87\textcolor{blue}{\textbf{(-1.93)}}\\
        \rowcolor{Gray}
        EA-7~\cite{shen2021efficient} & 28.28 & 1.473 & 0.77 &77.91\textcolor{blue}{\textbf{(-1.06)}}\\
        PA-4~\cite{choromanski2021rethinking} & 28.27 & 4.481 & 0.51 &78.73\textcolor{blue}{{(-3.07)}}\\
        \rowcolor{Gray}
        PA-7~\cite{choromanski2021rethinking} & 28.28 & 1.473 & 0.77 &77.87\textcolor{blue}{{(-1.1)}}\\
        AA-4~\cite{wu2021fastformer} & 28.27 & 4.394 & 0.50 &77.60\textcolor{blue}{{(-3.93)}}\\
        \rowcolor{Gray}
        AA-7~\cite{wu2021fastformer} & 28.28 & 1.445 & 0.75 &76.02\textcolor{blue}{{(-2.95)}}\\
        XCA-4~\cite{ali2021xcit}& 28.27 & 4.480 & 0.51 &78.67\textcolor{blue}{{(-3.13)}}\\ 
        \rowcolor{Gray}
        XCA-7~\cite{ali2021xcit}& 28.28 & 1.473 &0.77 &77.62\textcolor{blue}{{(-1.35)}}\\ 
        \midrule
        \multicolumn{5}{c}{References}\\
        \midrule
        Swin-4~\cite{liu2021swin}& 28.27 & 4.528 & 0.51 &80.08\textcolor{blue}{{(-1.72)}}\\
        \rowcolor{Gray}
        Swin-7~\cite{liu2021swin}& 28.28 & 1.500 & 0.78
        &78.72\textcolor{blue}{{(-0.25)}}\\
        LPI-4~\cite{ali2021xcit}& 28.38 & 4.520 & 0.51
        & 81.54\textcolor{blue}{{(-0.26)}}\\
        \rowcolor{Gray}
        LPI-7~\cite{ali2021xcit}& 28.39 & 1.486 & 0.78 &79.7\textcolor{red}{{(+0.73)}}\\
        COL-14& 22.00 & 6.117 & 0.69
        &81.30\textcolor{blue}{{(-0.50)}}\\
        COL-16& 22.00 & 4.589 & 0.52 &80.97\textcolor{blue}{{(-0.83)}}\\
        \bottomrule
      \end{tabular}}
      \caption{Experiment results with efficient attentions and benchmarks. COL means columnar structure and LPI means local patch interaction in \Cref{sec:eff}. The patch size is the number after the attention. The number parameters of the model and GFLOPs were measured with an input resolution of 224. Additionally, the ratio of FLOPs was calculated using a baseline of the same color.}
      \label{tab:result}
      \vspace{-2.5mm}
\end{table}

This section presents comparison experiments on the ImageNet1K dataset for the classification task. \Cref{tab:result} presents the number of parameters, FLOPs, and top-1 accuracy between various efficient attentions. \Cref{fig:result} shows the top-1 accuracy for various efficient attentions versus GFLOPs under different patch sizes. From \Cref{tab:result}, results show that no efficient attention performs better than the baseline. Furthermore, the performance order of each model with $4 \times 4$ patches is good in the order of EA, LA, FA, XCA, and AA. When the patch size is $7 \times 7$, the performance order is EA, PA, XCA, LA, and AA. In both cases, EA attained the least performance loss and AA achieved the most inferior accuracy with the largest reduction in computations. Also, PA, which uses a more complex kernel than the softmax kernel in EA, performed poorly. However, LPI showed similar or higher performance than the baseline. Compared to XCA, the performance of LPI was higher by 2.87\% and 2.08\% for $4\times 4$ and $7 \times 7$ patches, respectively. This shows that efficient attention can outperform the normal self-attention when used with a proper module.

\section{Conclusion}
In this paper, we conduct a comparison experiment between pyramid transformers with efficient attentions. For a fair comparison, we considered the same environments for all experiments. The experimental results show that efficient attention achieved lower accuracy and computation than normal self-attention. However, in some cases, efficient attentions perform on par or better than the baseline. This result shows the potential power of efficient attentions to reduce computation. 

\textbf{Acknowledgments} This work was partially supported by a National Research Foundation of Korea (NRF) grant funded by the Korean government (MSIT) (No. 2020R1C1C1007423 and project BK21 FOUR).
\clearpage
{\small
\bibliographystyle{ieee_fullname}
\bibliography{main}
}

\end{document}